\documentclass[a4paper,10pt]{article}
\usepackage{a4wide}

\usepackage{graphicx}
\usepackage{tabu}
\usepackage{multirow}

\usepackage{hyperref}
\usepackage{authblk}

\makeatletter
\def\blfootnote{\xdef\@thefnmark{}\@footnotetext}
\makeatother

\title{Human-centred explanation of rule-based decision-making systems in the legal domain}

\author[A]{Suzan Zuurmond}
\author[B]{AnneMarie Borg}
\author[A]{Matthijs van Kempen}
\author[A]{Remi Wieten}
\affil[A]{Dutch Tax and Customs Administration, The Netherlands}
\affil[B]{Utrecht University, The Netherlands}
\date{}

\begin{document}
\maketitle

\begin{abstract} 
We propose a human-centred explanation method for rule-based automated decision-making systems in the legal domain. Firstly, we establish a conceptual framework for developing explanation methods, representing its key internal components (content, communication and adaptation) and external dependencies (decision-making system, human recipient and domain). Secondly, we propose an explanation method that uses a graph database to enable question-driven explanations and multimedia display. This way, we can tailor the explanation to the user. Finally, we show how our conceptual framework is applicable to a real-world scenario at the Dutch Tax and Customs Administration and implement our explanation method for this scenario.
\end{abstract}

\section*{Introduction}
\label{sec:Introduction}
In the mid-eighties, governments worldwide started using automated decision-making systems to improve the efficiency and effectiveness of public administration~\cite{TimmerR19rule-based}. Often, these systems are so-called rule-based systems, with rules based on legislation. These systems have advantages over human workers since they are consistent, cost-efficient and often expedite the decision-making. However, they can also impact individuals' rights or legal quality~\cite{Eck18geautomatiseerde}, especially when there is no careful and transparent design process~\cite{Lokin18wendbaar}. \blfootnote{This is the full version of a demo at the 36th International Conference on Legal Knowledge and Information Systems (JURIX'23).}

As the ongoing digitisation of governance raises concerns globally, there is a pressing need for innovation and digital technologies that improve transparency~\cite{OpenGovernmentPartnership20}. A transparent decision-making process provides all the relevant information on the input data, system design and processes and allows individuals to understand this information as well as the system's decisions~\cite{IHLEGAI19}. 

An example of an innovation that improves the transparency of legislative automation is the Agile Law Execution Factory (ALEF)~\cite{JetBrains19}, in use at the Dutch Tax and Customs Administration (DTCA). ALEF is a tool for developing and generating service applications that can perform specific legal tasks like tax calculations. It uses a domain-specific language called Regelspraak~\cite{Regelspraak23} to allow for the specification of rules, data descriptions and test cases, which are interpretable for legal experts and developers while being executable for the software system itself~\cite{CorsiusHLBSW21}. Using Regelspraak, ALEF combines legislative and policy analysis descriptions in interpretable knowledge representations. While tools like ALEF could contribute to explaining automated decisions~\cite{LokinK19}, they often lack adequate explanation mechanisms.

In this paper, we introduce a human-centred explanation method for rule-based decision-making systems within the legal domain. As a result, (i) we improve the transparency of the design process by bridging the gap between the transparency required by regulations and the complexity of practical implementation, and (ii) we introduce an explanation mechanism for systems, like the ones created using ALEF, offering a way for generating and conveying explanations that are tailored to the needs of individuals involved with these systems generated by such tools.

First, in Section~\ref{sec:ConceptFramework}, we propose a conceptual framework for explanation methods for automated decision-making systems, based on internal components~\cite{LacaveD02} 
and external dependencies~\cite{SormoCA05}. Then, in Section~\ref{sec:DTCA}, we show how this framework can be applied to the context of ALEF and the DTCA. 

Subsequently, in Section~\ref{sec:Method}, we develop an explanation method that integrates graph databases with question-based explanations. This approach ensures flexibility within the internal components and adaptability concerning the external dependencies. Furthermore, we offer a variety of explanation formats, including textual, tabular, and visual representations.

To illustrate the practical application of our explanation method, we conducted a case study within the DTCA in Section~\ref{sec:Study}, focusing on the tax interest calculation system developed using the ALEF tool. 
We show how our explanation method can be used to generate explanations tailored to different target audiences. As a result, we improve the explainability of automated decisions made at the DTCA and ensure that both the decision-making and the applied legislation become transparent to the individuals involved with the system



\section{Conceptual Framework for Explanation Methods}
\label{sec:ConceptFramework}
\begin{figure}[h!]
    \centering
    \includegraphics[width=0.8\textwidth]{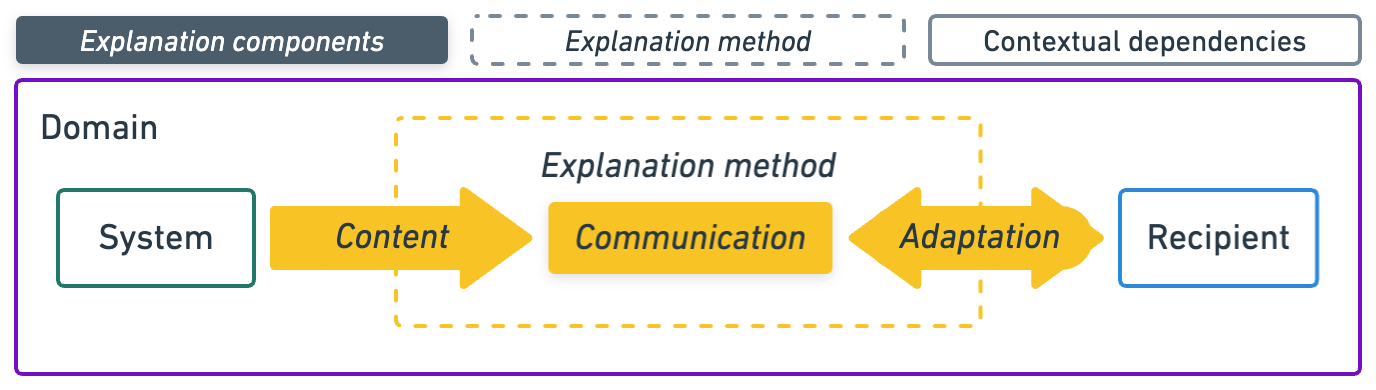}
    \caption{Overview of the components and dependencies of an explanation method.}
    \label{fig:conceptual_framework}
\end{figure}

In this section, we introduce a conceptual framework illustrating an explanation methods' internal components and external dependencies (see Figure \ref{fig:conceptual_framework}). This framework will serve as a guide for this research.

In the literature, various analyses reveal three common aspects of explanation: identifying relevant information and constructing justifications; conveying explanations effectively; and tailoring explanations to the target audience's needs~\cite{Miller19, NeerincxWKD18, LacaveD02}. These components collectively form the core of explanation methods and focus on the following questions, named after \cite{LacaveD02}:

\begin{description}
    \item[\textbf{Content}] \textit{what} should be included in the explanation?
    \item[\textbf{Communication}] \textit{how} should the explanation be conveyed?
    \item[\textbf{Adaptation}] to \textit{whom} and how should the explanation be tailored?
\end{description}

The effectiveness of an explanation method, as well as the explanations it produces, are highly contextual ~\cite{Miller19}. More specifically, the objectives an explanation needs to fulfil can differ greatly based on the domain, the characteristics of the AI system, and the users involved ~\cite{SormoCA05}. So, when developing an explanation method, one should consider the following external dependencies:

\begin{description} 
    \item[\textbf{Domain}] the operational context in which the system operates;   
    \item[\textbf{System}] properties as complexity, input or output type, and autonomy level;
    \item[\textbf{Recipient}] the goals and knowledge of the individuals receiving explanations.
\end{description}

This distinction of dependencies is represented in our primary objective: to craft a \textit{human-centred} (recipient) explanation method for a \textit{rule-based decision-making system} (system) in the \textit{legal domain} (domain).

\section{Automated decision-making systems at the DTCA}
\label{sec:DTCA}
In this section, we leverage the conceptual framework to explore the external dependencies within the context of the Dutch Tax and Customs Administration, particularly focusing on decision-making systems developed using the ALEF tool. 

\paragraph{Domain. }
The legal domain is complex, with continuously evolving laws and intricate regulations. Therefore, the DTCA employs Law Analysis, outlined in~\cite{LokinAB21}, ensuring clarity in translating legislation into practice, enabling justifiable decisions, and facilitating adaptability to changing legislation in ICT systems.

\begin{figure}[h!]
    \centering
    \includegraphics[width=0.8\textwidth]{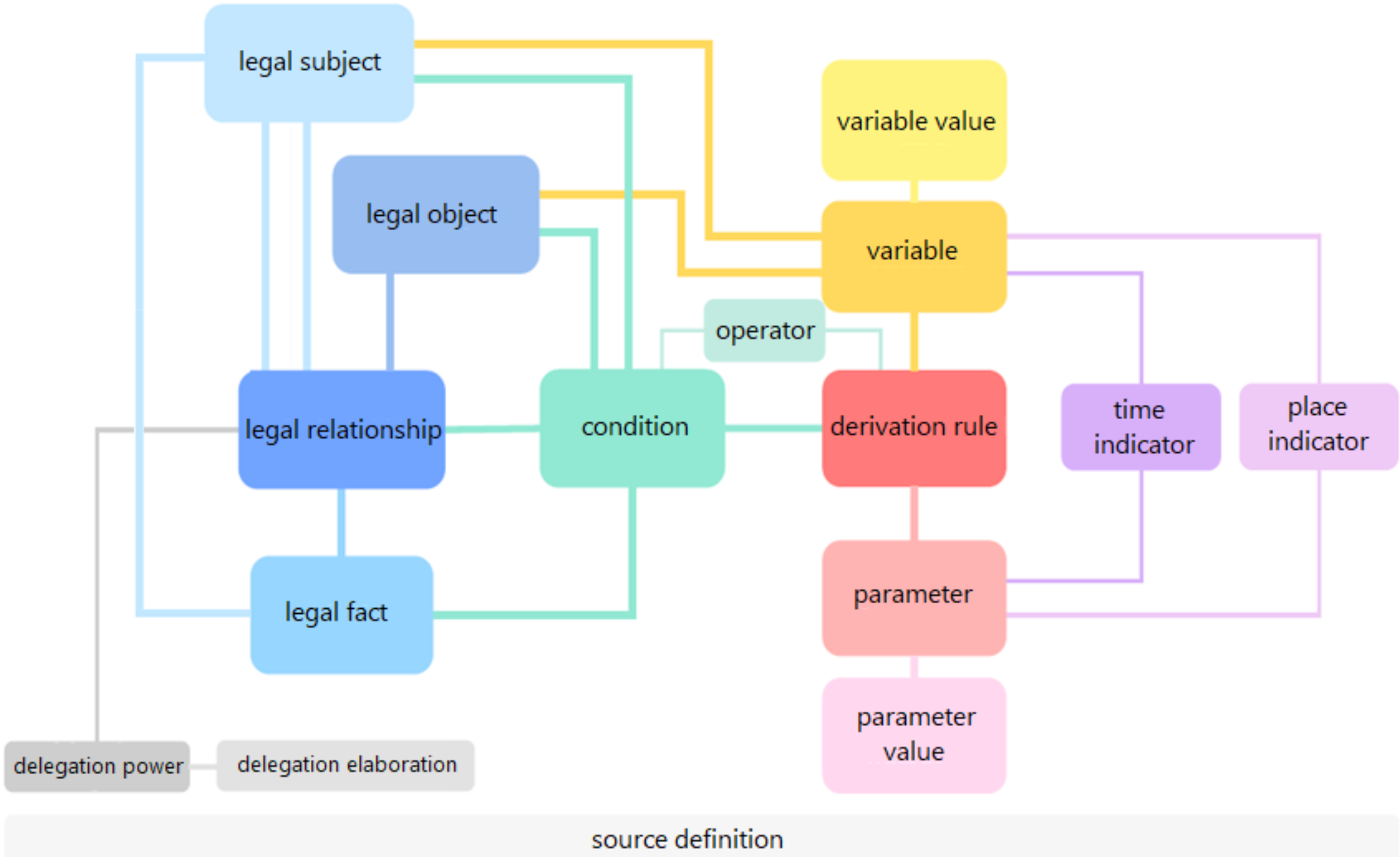}
    \caption{Diagram representing a framework for legal analysis, translated from~\cite{LokinAB21}.}
    \label{fig:legal_analysis}
\end{figure}

A key element of this approach is the legal analysis schema, drawing from~\cite{Hohfeld13, Hohfeld17}, offering a visual overview of legislative components and relationships (see Figure~\ref{fig:legal_analysis}). This schema identifies legal concepts like legal subject, object, relationship, and fact, which are vital for understanding and applying legislation. It also includes conditions and derivation rules. As these concepts are the building blocks for constructing models used in decision-making systems, this schema will be used to structure the explanations in Section~\ref{sec:Method:Implementation}.

\paragraph{System. }
\label{sec:DTCA:System}

The decision-making systems generated by the ALEF tool~\cite{JetBrains19} function through a rule-based reasoning process, utilising pre-defined rules that are explicitly programmed into the systems. These rules typically take the form of (propositional) if-then statements with distinct condition and action components. For example, \textbf{if} the tax payment is not made by the specified due date, \textbf{then} tax interest is applied to the unpaid amount over time. When employing this rule within a system, once the condition of missing the specified due date for payment is satisfied, the corresponding action entails the imposition of tax interest.

We focus on creating automated explanations to be applicable and practical in real-world settings like the DTCA. This means the method should generate explanations with minimal human intervention after the initial setup. Additional factors, such as efficiency and scalability, are beyond the scope of this paper.

While the case study (Section~\ref{sec:Study}) is aimed at explanations for decisions derived by the ALEF tool, the explanation method is based on more generic characteristics. This ensures that our method can also be applied to other rule-based decision-making systems. 

\paragraph{Recipient. }
\label{sec:DTCA:Recipient}
Individuals ask questions because they have a particular goal, such as understanding a system's behaviour or accomplishing specific tasks. A question-driven framework, aided by visualisation, assists in connecting these inquiries with their fundamental goals and provides an adaptable and effective explanation method~\cite{HoffmanMKL19,LimD09}.

We adapt the taxonomy of user needs from~\cite{LimD09} to rule-based systems and the legal domain. Table~\ref{tab:type_questions} presents the relevant explanation types for automated legal decisions, divided into two categories. The \emph{system} category includes global explanations on the general decision model, while the \emph{decision} category includes local explanations regarding a specific decision-making instance. Moreover, we differentiate the purpose of explanation types based on whether they describe something or clarify something~\cite{LacaveD02}. 

Compared to~\cite{LimD09}, we add the \emph{whether} type to provide information on whether certain requirements are met. The need for this question type builds on the necessity of verification (see Section~\ref{sec:Study:model}). Furthermore, we distinguish between \emph{how} and \emph{how to} explanations, where the first clarifies how the system generally makes certain (intermediate) decisions, while the latter clarifies how to achieve a desired decision, given some (but not all necessary) input values. This improves the adaptability of our explanations. 

\begin{small}
    \begin{table}
    \begin{tabu} to \textwidth  {| X[.1,c] | X[.4,l] | X[2,l] |}
    \hline 
     & {\bf Type} & {\bf Purpose} \\
    \hline
     \multirow{7}{*}{\rotatebox[origin=c]{90}{Decision}} 
    & What      & Describes what decision(s) it made and what input it used. \\
    & What if   & Describes the system’s decision when an input value changes.\\
    & Why       & Clarifies the reasoning behind the system’s decision, given the input values.\\
    & Why not   & Clarifies why an alternative decision was not generated, given the input values.\\
    & How to    & Clarifies how to achieve a desired decision, given some (but not all necessary) input values.\\
    \hline
    \multirow{5}{*}{\rotatebox[origin=c]{90}{System}} 
    & Input     & Describes which input the system could use.\\
    & Output    & Describes which decisions the system could make.\\
    & How       & Clarifies how the system generally makes a certain decision.\\
    & Visualisation & Describes a simplified view of the system's conceptual model.\\
    & Whether & Clarifies whether the system meets a certain requirement.\\
    \hline
    \end{tabu}
    \caption{Different explanation type questions for automated legal decisions, building on the explanation types in~\cite{LimD09}.}
    \label{tab:type_questions}
    \end{table}
\end{small}

This set of explanation types ensures that our explanation method effectively addresses the unique goals and information needs of all individuals involved in rule-based decision-making systems in the legal domain.

\section{Explanation Method}
\label{sec:Method}
We can now present a human-centred explanation method for rule-based decision-making systems in the legal domain. We first discuss using Graph Database Management Systems and then provide a short implementation overview.  

\subsection{Explanation through Graph Database Management Systems}
\label{sec:Method:GDBMS}

We propose an explanation method using Graph Database Management Systems (GDBMS), specifically Neo4j~\cite{Lal15}. These make it possible to store data in graph structures and allow for intuitive questioning of relational data and visualising relationships between data points. 
%

In our explanation method, we utilise GDBMS to structure our explanations according to the legal analysis schema (Figure~\ref{fig:legal_analysis}) and we combine it with question-driven user-system interaction and multimedia display to account for the three internal components. More specific, in GDBMS, nodes can, for example, represent legal subjects, legal objects, requirements, rules and variables, while the connections represent their legal relationships. Recipients can question both the decision model and a specific decision (focus), filter information by, for instance, focusing on a specific part, node or relationship (level of detail) and extract causal relations between conditions, rules and derivations of these rules (causality), thereby varying the content of the explanation. Explanations are answers to predefined questions that can be communicated through visual graphs, verbal text answers and customised filtering options. Finally, GDBMS enables the creation of different perspectives for different recipient categories, thereby adapting the explanation to the recipient, by assuming a certain level of knowledge (both domain and reasoning knowledge) and linking a specified goal to a set of specific questions. 

\subsection{Implementation of the Explanation Method}
\label{sec:Method:Implementation}

Due to space constraints, we only provide a high-level sketch of the implementation.\footnote{For the full implementation, see \url{https://github.com/sjzuurmond/mps_explaineo}.} We start by importing all the model elements in a structured form, creating an abstract syntax graph containing the knowledge from all the decision models. From this graph, we create a simplified graph representing the decision models’ key structural and semantic information. This simplification is based on the goals and knowledge of the recipient of the explanation (e.g., in  our case study in Section~\ref{sec:Study} we used the legal analysis schema from Figure~\ref{fig:legal_analysis}). To represent a specific decision, we can instantiate the simplified graph with the values of a model instance.

\paragraph{Object, Rule and Service Model. } In practice, a decision system uses multiple models representing its reasoning. ALEF uses three models: the Object Model, representing the entities and their attributes involved in decision-making; the Rule Model, representing rules and their dependencies for making decisions; and the Service Model, representing services used by legal professionals for decision-making by providing input and output variables. 

Variables play a central role in the decision-making process and are the different models’ overlapping factors. Hence, a clear representation of variables aids in understanding the decision-making process. For example, variables represent the attributes of objects in the Object Model, the input and output fields of services in the Service Model are mapped onto variables, and the rules in the Rule Model use variables in their conditions or calculations or to derive new variable values.

The Object Model consists of variables and object types with ``relates\_to" relationships between object types. In the decision-making system, a distinction is made between Boolean features and attributes of objects. However, in line with the diagram, we will refer to them as variables. The Service Model consists of input and output messages with input and output relationships to the connected variables. Finally, the Rule Model consists of rules with calculation, derivation, and condition relationships from and toward the corresponding variables.

\paragraph{Relation Between Legal Concepts and Models. } The legal concepts in the legal analysis diagram (Figure~\ref{fig:legal_analysis}) can be related to the elements in the models used by the decision-making system. For example, the Object Model represents legal subjects, objects, and their relationships. Legal subjects and objects correspond to the object nodes in the Object Model, while the connections between these nodes can represent legal relationships. The Rule Model can represent conditions, derivation rules, and requirements that must be fulfilled. In addition, operators can also be represented as rules in the Rule Model. Rule nodes correspond to specific rules or conditions, while variable-type nodes in the Rule Model represent the variables these rules use. At last, the Service Model can represent services and their input and output variables. Service nodes correspond to specific services used during the decision-making process, while variable type nodes in the Service Model represent input and output variables for these services.

Note that not all the legal concepts are represented and captured in the models, e.g.\ delegation power and delegation elaboration. However, the modeller can add other relevant information. For example, they can set the sources of rules. This added information is extracted from the model elements and represented as properties in the summarising graph.

\section{Case Study}
\label{sec:Study}

We concentrate on an application within the domain of tax law – the tax interest calculation system. This application plays a crucial role in determining interest rates for tax assessments in the Dutch tax landscape, where tax professionals rely on it to calculate rates with significant financial implications for taxpayers. Therefore, explanations are essential for all users involved; we consider two.

First, model experts implement tax regulations into decision models. These experts are responsible for ensuring that the application accurately reflects the complex web of tax laws and regulations. Clear explanations not only aid in the model's development but also enable experts to verify that the system aligns with legal requirements, promoting regulatory compliance. Second, legal support professionals bridge the gap between complex tax decisions and taxpayers seeking clarity. When communicating with taxpayers, they require a comprehensive understanding of the underlying decision-making process to provide concise and accurate explanations. Thus, explanations serve as a vital contribution, fostering effective help for model experts and legal support professionals by promoting regulatory adherence and ensuring taxpayers receive the clarity they deserve.

We show how our method creates explanations for these two target groups. 

\subsection{Model Experts}
\label{sec:Study:model}

The model expert is involved in the creation and verification of the model. When creating the model, our explanation method offers guidance in this modelling process by providing a visual representation of each model to aid the textual editor of ALEF. Such a visualisation contains the variables, objects, rules and input/output messages as nodes as well as the relationships between these nodes. What type of nodes and relations are included depends on the type of question. For example, when asking \emph{What elements has the Object Model?} the nodes are the variables and objects, while when asking \emph{What elements has the Rule Model?} there are nodes for rules as well. 

The verification process for a rule-based decision-making system is essential to ensure its proper functioning. We distinguish three types of checks to guide the verification process:
\begin{description}
    \item[Path checks] verify if the system utilises an element and help identify redundant elements.
    \item[Assignment checks] ensure an element is assigned a value, confirming that these elements can provide results.
    \item[Logical checks] confirm a rule's absence of logical contradictions in its conditions and help identify any inconsistencies.  
\end{description}

This distinction generalises the verification documentation as drafted by model experts at the DTCA for the ALEF tool used in the case study, but are generally applicable in decision-making systems. The same documentation states several questions a modeller should answer themselves to verify the models they made: 
\begin{itemize}
    \item \emph{Is each in and output message used by the Service?} (path check)
    \item \emph{Is all input used to create the output?} and \emph{Can all output be created, given the input?} (path check)
    \item \emph{Is each variable used?} (path check)
    \item \emph{Are all variables assigned?} (assignment check)
\end{itemize}
The answers to these questions contain (i) a text message stating whether the answer to the question is positive or negative; (ii) a table with for each message or variable whether it is used or a path has been found; and (iii) a visualisation of the textual answers, similar to those for model creation. 

For example, in order to understand the created system better, the model expert can ask \emph{Can all output be created?}. If this is not the case, the system shows which variables cannot be derived (e.g., the start and end date of the interest period), in text, in a table it shows all the variables and whether there is a path and the visualisation allows the expert to understand the relations and see how the variables could be assigned.\footnote{The GitHub page \url{https://github.com/sjzuurmond/mps_explaineo} contains images of the three types of answers.} 

\subsection{Legal Support Professionals}
\label{sec:Study:support}

Legal support professionals work within customer service or other relevant departments to provide legal support and explanation to data subjects regarding the decisions made. Hence, they need to be able to answer relevant questions for a data subject. For these questions, a distinction is made between two types of explanation (matching two of the explanation types made by ICO~\cite{InformationCommissionersOffice}):
\begin{description}
    \item[Data explanation] explaining the data used in the decision-making system. 
    \item[Rationale explanation] explaining the reasoning behind a decision made by a decision-making system. 
\end{description}
A data explanation answers: \emph{What decisions did the system make?} by providing a table with the output message, variable and assigned values; \emph{What (personal) information was used for these decisions?} by providing a table with the input message, variable and given values; and \emph{Which rules were used for these decisions?} by giving all rules used in the decisions. A rationale explanation answers: \emph{Why is this decision made?} by providing a textual description of the rule and conditions that are met, aided by a visualisation which shows the variable representing the decision, the rule that determined this variable and the derivation relationships from this rule to the variable as well as the conditional variables and relationships; and \emph{Why is this decision made? (trace)} by providing a textual description of the rules and conditions that are met like a trace and again aided by a visualisation. 

\begin{figure}
    \centering
    \includegraphics[width=\textwidth]{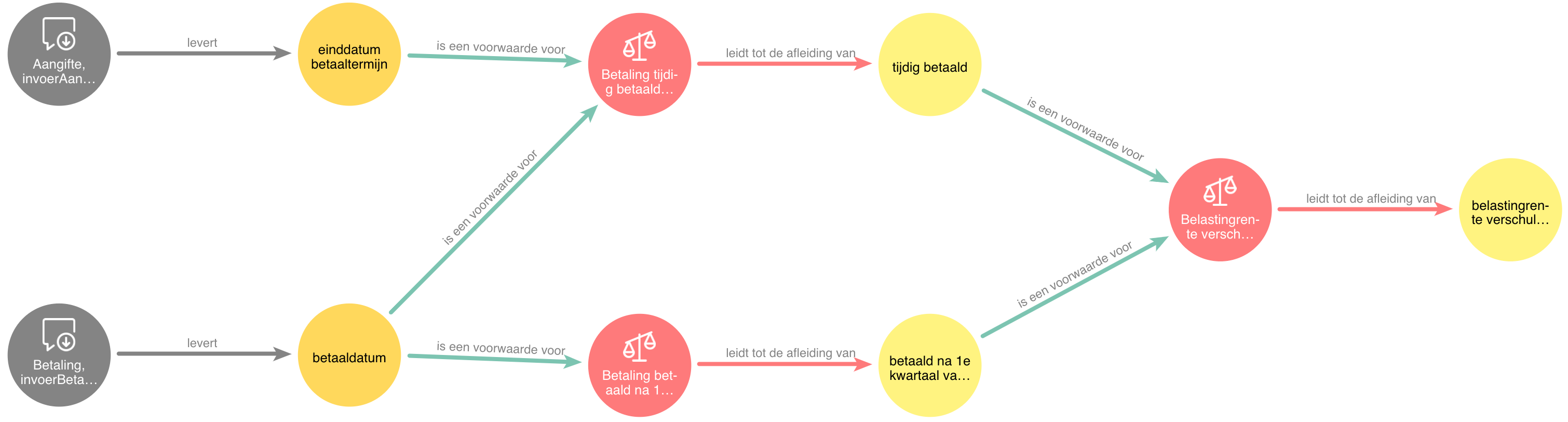}
    \caption{Visualisation of the explanation for \emph{why do they have to pay tax interest?}, with a trace from input messages to decision.}
    \label{fig:trace}
\end{figure}

For example, when a taxpayer asks a legal support professional why they have to pay tax interest, the system shows the professional, in text, the applicable rules, including a link to the specific law (e.g., taxpayer paid taxes too late, \href{https://wetten.overheid.nl/jci1.3:c:BWBR0002320&hoofdstuk=VA&artikel=30h&z=2023-01-01&g=2023-01-01}{link}), and the conditions that are satisfied (e.g., the required payment specification was made at a specific date and at least one payment was overdue) as well as a visualisation of the conclusion, the rules and conditions and their relations, see Figure~\ref{fig:trace}.\footnote{See the GitHub page \url{https://github.com/sjzuurmond/mps_explaineo} for the visualisation as well as the trace variant of the explanation.} The professional can then communicate this with the taxpayer.  

\section{Discussion and Conclusion}
\label{sec:Conclusion}

In this paper, we have proposed a framework for developing explanation methods that offer a human-centred approach to explaining rule-based decision-making systems in the legal domain. By focusing on content, communication, and adaptation, the method ensures that users can obtain the information they need to understand the decision-making process and trust its decisions. Using graph databases further enhances the explanatory value of the method, facilitating more effective communication and adaptation to certain individuals' needs. Since our explanation method is connected to the legal analysis diagram, explanations provide insight into the decision-making system and the legislation. We also implemented the proposed method in a real-world scenario: an automated decision-making system used by the DTCA. 

While our primary focus has been on DTCA and ALEF, our explanation method holds significant value for a broader spectrum of automated decision-making systems. For instance, the conceptual framework introduced in Section~\ref{sec:ConceptFramework} offers a versatile foundation applicable to a wide range of explanation methods. Moreover, our approach successfully bridges the gap between well-established explanation methodologies~\cite{LacaveD02, SormoCA05, Miller19, NeerincxWKD18} and cutting-edge technologies~\cite{JetBrains19, Neo4j}. The integration of graph databases, combined with question-driven user-system interaction and multimedia display, exemplifies how our method harnesses both traditional principles of human explanation and emerging technologies to craft more effective and human-friendly explanations.


Our explanation method lacks the ability to reason independently. That is, other than revealing relations, the graph database lacks the capability to re-calculate or reason about decisions independently. Consequently, for the model expert, it can answer questions related to path and assignment checks, but it cannot perform logical checks since it cannot compare two conditions. Similarly, for the legal professional, our explanation method can answer `what' and `why' questions, but it cannot provide answers to `what if' and `why not' questions. Extending the explanation method to include reasoning capacity or a live integration with the decision system is part of future work. 

There are many possible directions for future work, from extending our explanation method to allow for, e.g., dynamics in the target group and other explanations forms, to improving the visualisation. Additionally, implementing the explanation method for the DTCA provides a good starting point for applying the method to other domains. We can then further explore the flexibility of the method and its ability to cater to the unique needs of different user groups. 



\bibliographystyle{vancouver}
\bibliography{literature}

\end{document}